\documentclass{article}
\usepackage{spconf,amsmath,graphicx,balance}
\usepackage{amsfonts}
\usepackage{hyperref}
\usepackage{multirow}
\usepackage{setspace}
\usepackage{caption}
\usepackage{subcaption}
\usepackage{xcolor}
\usepackage{soul}
\usepackage{changes}
\definechangesauthor[name={Amil Dravid}, color=violet]{ad}

\usepackage{changes}
\definechangesauthor[name={Florian Schiffers}, color=orange]{fs}

\hypersetup{
    colorlinks=true,
    linkcolor=black,
    filecolor=black,
    citecolor=black,
    urlcolor=blue,
}

\title{Investigating the Potential of Auxiliary-Classifier GANs for Image Classification in Low Data Regimes}
\makeatletter

\name{Amil Dravid$^{*1}$, Florian Schiffers$^1$, Yunan Wu$^2$, Oliver Cossairt$^{1,2}$, Aggelos K. Katsaggelos$^{1,2}$}
\address{
$^1$Department of Computer Science, Northwestern University, Evanston, IL 60208, USA \\
$^2$Department of Electrical and Computer Engineering, Northwestern University, Evanston, IL 60208, USA\\
{$^*$amildravid2023@u.northwestern.edu}} 

\begin{document}
%
\maketitle
\begin{abstract}
  Generative Adversarial Networks (GANs) have shown promise in augmenting datasets and boosting convolutional neural networks' (CNN) performance on image classification tasks. But they introduce more hyperparameters to tune as well as the need for additional time and computational power to train supplementary to the CNN. In this work, we examine the potential for Auxiliary-Classifier GANs (AC-GANs) as a ’one-stop-shop’ architecture for image classification, particularly in low data regimes. Additionally, we explore modifications to the typical AC-GAN framework, changing the generator’s latent space sampling scheme and employing a Wasserstein loss with gradient penalty to stabilize the simultaneous training of image synthesis and classification. Through experiments on images of varying resolutions and complexity, we demonstrate that AC-GANs show promise in image classification, achieving competitive performance with standard CNNs. These methods can be employed as an ’all-in-one’ framework with particular utility in the absence of large amounts of training data.


 %
 

\end{abstract}
\begin{keywords}
Generative Adversarial Networks, Image Classification, Convolutional Neural Networks, Data Augmentation, Deep Learning
\end{keywords}
\section{Introduction}
\label{sec:intro}

Convolutional Neural Networks (CNNs) are widely used for image classification in applications from natural image classification to computer-aided diagnosis of illnesses~\cite{pan2019recent}. 
%
However, they rely on large training datasets in order to generalize well on validation or testing datasets~\cite{wang2018transferring}. Furthermore, obtaining more data can be costly and time prohibitive. 
%
%
%
%
Even in the case of models trained on large datasets, performance often drops when tested on similar data from other sources or collected in a different manner~\cite{recht2018cifar}.
%
%
%

Generative Adversarial Networks (GANs)~\cite{goodfellow2014generative} provide one avenue to remedy the data scarcity problem. They show particular promise in data augmentation given their ability to produce new images mimicking a prior collection~\cite{shin2018medical}. A GAN consists of a \emph{discriminator} and a  \emph {generator} model that are trained in tandem.
%
%
The generator $G$ carries out \emph{implicit density estimation}, whereby it learns a function to sample from an estimated probability distribution $p_g$ and generate data $G(z)$ that mimics the true data distribution $p_{data}({\emph{x}})$. 
Furthermore, the generator tries to map onto $p_{g}$ a \emph{latent} or \emph{noise} vector $z$ drawn from a simple distribution $p_{\textbf{\emph{z}}}$, such as a normal distribution. The generator is trained to minimize the divergence between $p_{g}$ and $p_{data}$, whereas the discriminator tries to maximize the divergence. This entire process can result in photo-realistic images or accurate data of other modalities~\cite{pan2019recent}. Thus, GANs can provide the much-needed additional images for CNNs. However, this entails training two separate models: the CNN for classification and the GAN for augmentation. \\
\indent  The Auxiliary-Classifier GAN (AC-GAN)~\cite{odena2017conditional} builds upon the standard GAN. Inspired by the class-conditioned GAN (C-GAN)~\cite{mirza2014conditional}, the generator is fed class information as a condition via a one-hot-encoded vector concatenated to the noise vector. The discriminator then outputs both the source of the inputted image (real or fake) and a second label corresponding to the input's class. This enables the generator to synthesize images with greater fidelity to the class label. 
The AC-GAN follows a two-part objective function:
\begin{equation}
\footnotesize
{L_S = \mathbb{E}[\log P(S = real | X_{real}))] + \mathbb{E}[\log P(S = fake | X_{fake})]}
\label{eqn:3}
\end{equation}
\begin{equation}
\footnotesize{
L_C = \mathbb{E}[\log P(C = c | X_{real}))] + \mathbb{E}[\log P(C = c | X_{fake})] },
\label{eqn:4}
\end{equation}

%
%
%
%
where $L_S$ denotes the log-likelihood of the discriminator assigning the correct source $S$, real or fake, and $L_C$ denotes the log-likelihood of the discriminator assigning the correct class $C$ given real and fake images.
The discriminator seeks to maximize $L_S + L_C$, while the generator tries to maximize $L_C - L_S$.

The AC-GAN objective was constructed to encourage the generator to produce more class-discriminable images \cite{odena2017conditional}. Yet, there is a gap in knowledge regarding the auxiliary classifier's performance. Our contribution lies in examining the effectiveness of the AC-GAN for image classification in multiple domains, from binary to multi-class classification, and small to large-scale images. Motivated by GANs' utility in augmenting data, we particularly investigate classification using small-sized datasets. Furthermore, we propose simple modifications to the AC-GAN training scheme to facilitate classification. This can serve as an all-in-one framework that avoids the need to train a separate GAN and CNN. 

\section{Related Works}
\label{sec:relatedworks}
%
%
%
%

%
%
%
The literature on AC-GAN's potential as an image classifier is sparse. \cite{frid2018gan} employs AC-GANs for improving a 3-class CNN liver lesion classifier.
One of the experiments entails using an isolated discriminator from an AC-GAN for classification, which realizes a ${\sim}{2}$\% decrease in performance in comparison to their CNN model. 
However, they employ the exact same original AC-GAN architecture presented in~\cite{odena2017conditional}, whereas their CNN classifier is constructed with the explicit objective of accurate liver classification.
%
%
%
%
Yet, all-in-one architectures that combine generation and classification have shown promise in the literature~\cite{vandenhende2019three,9173809, wang2019early}.
For instance, \cite{wang2019early} proposes a domain-specific variant of the AC-GAN for effective spectrum classification in hyperspectral imaging, plant segmentation, as well as image classification. Additionally, the work in~\cite{zhao2018application} applies an Auxiliary-Classifier Wasserstein GAN with gradient clipping to a specific signal classification task. These works provide motivation to understand, in general, how effective AC-GANs can be as an image classifier when compared to standard CNNs in a controlled setting (i.e., similar hyperparameters, adapted to the same domain/task, etc.). We explore methods to address this and further adapt AC-GANs to image classification, particularly in low data regimes. 
%
%

\vspace{-3mm}

\section{Method}

\label{sec:method}
\subsection{Wasserstein loss with gradient penalty}
\label{ssec:ac-gan_loss}
GAN training can often be unstable as the generator may ultimately only produce low-fidelity images or produce a single output that fools the discriminator, known as \emph{mode collapse}~\cite{che2016mode}.
Traditionally, the binary cross-entropy function is used as a loss function to approximate the \textit{Jensen-Shannon} (JS) divergence between the generator's estimated density model $p_g$ and the underlying data distribution $p_{data}$~\cite{goodfellow2014generative, goodfellow2016nips}.
The discriminator seeks to maximize this divergence. 
This may lead to saturated gradients, and little valuable feedback for the generator to improve~\cite{arjovsky2017wasserstein, yi2019generative}.
A \textit{Wasserstein} loss \cite{arjovsky2017wasserstein} approximates the \textit{Earth Mover's Distance} (EMD) between the two distributions, which has shown to mitigate these issues. 
However, the discriminator's gradients need to be \textit{1-Lipschitz} continuous in order to have a valid approximation of the EMD. That is, the norm of the discriminator's gradients must be at most one. A gradient penalty scheme~\cite{gulrajani2017improved} encourages this constraint, improving upon clipping the weights as proposed in~\cite{arjovsky2017wasserstein}. 

We employ the Wasserstein Loss with gradient penalty to model the loss for the discriminator assigning the correct source to an image :$L_S$. We keep Equation~\ref{eqn:4} as is, modeling the log-likelihood of the class. However, we modify $L_S$ to be: 
\begin{equation}
    L_S = \mathbb{E}[D(\emph{x})] - \mathbb{E}[D(G(\emph{z}))]
\end{equation}
where $D$ represents the discriminator's output for the source, $\emph{x}$ represents true data, and $G(\emph{z})$ denotes synthesized samples from the generator. The generator seeks to maximize $L_C-L_S$, while the discriminator tries to minimize $L_S-\omega L_C+\lambda\Phi$. The terms $\lambda$ and $\Phi$ represent a gradient penalty coefficient and regularization term to encourage 1-Lipschitz Continuity. The method for calculating the regularization term is further explained in~\cite{gulrajani2017improved}. We also introduce a weight $\omega$ for the class loss component in order to balance the discriminator's focus between classifying images as well as identifying the source. Introducing a Wasserstein loss with gradient penalty allows us to train an AC-GAN for longer to optimize its classification without sacrificing the generator's training. 


\subsection{Sampling the latent vector $z$ }
\label{ssec:sampling_z}
Feeding in poorly generated examples into a classifier may harm classification performance, so it is ideal to augment with the most realistic possible data~\cite{wang2019early, tanaka2019data}. To that end, in training the AC-GAN's discriminator for image classification, we propose using the so-called \emph{truncation trick} presented in \cite{brock2018}. Oftentimes, images from a generator are produced by sampling the latent vector $z$ from the spherical normal distribution $\mathcal{N}(0, \emph{\textbf{I}})$.
The truncation trick entails sampling closer to the mode of the distribution (also the mean in $\mathcal{N}(0, \emph{\textbf{I}})$), resulting in images with greater realism, but low diversity. On the contrary, sampling further away from the mode results in lower fidelity images but with greater diversity~\cite{brock2018}. Figure 1 provides an example to illustrate this concept.
%
%
%

We suggest to feed in generated images sampled by a truncated normal distribution when optimizing $L_C$ for the discriminator. That is, sampling within a specific domain that is closer to the mode.
Note that when optimizing the discriminator's parameters with respect to $L_C$ as well as all the generator's parameters, we sample from a standard $\mathcal{N}(0, \emph{\textbf{I}})$. This facilitates a standard GAN training scheme that will not result in significant trade-off between diversity and fidelity in the generator. However, the truncation sampling technique is applied to optimize $L_C$ with respect to the discriminator's objective, which will feed in higher fidelity images to improve the discriminator's image classification task. Our experiments prove this to be a useful and novel adaptation of the truncation trick, which has previously only been used after training for visualizing the generator's performance. 
\begin{figure}
  
  \centering

     \begin{subfigure}[b]{0.49\linewidth}
         \centering
         \includegraphics[width=\linewidth]{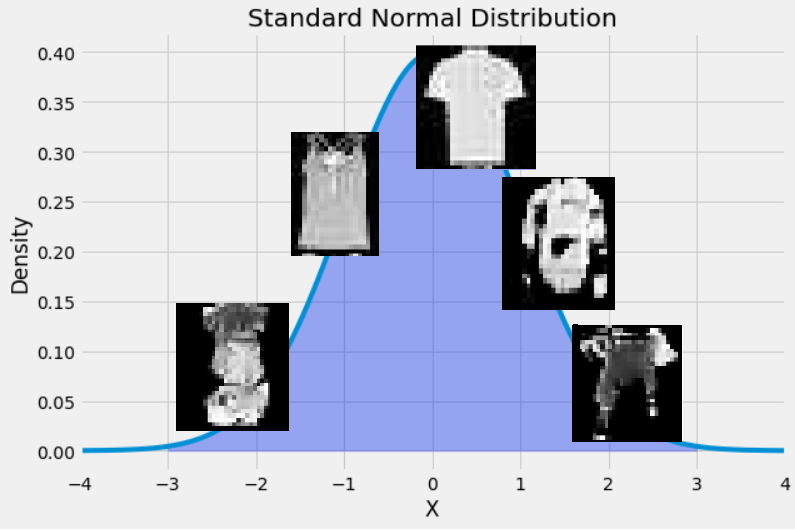}
         \caption{Full distribution}
     \end{subfigure}
     \hfill
     \begin{subfigure}[b]{0.49\linewidth}
         \centering
         \includegraphics[width=\linewidth]{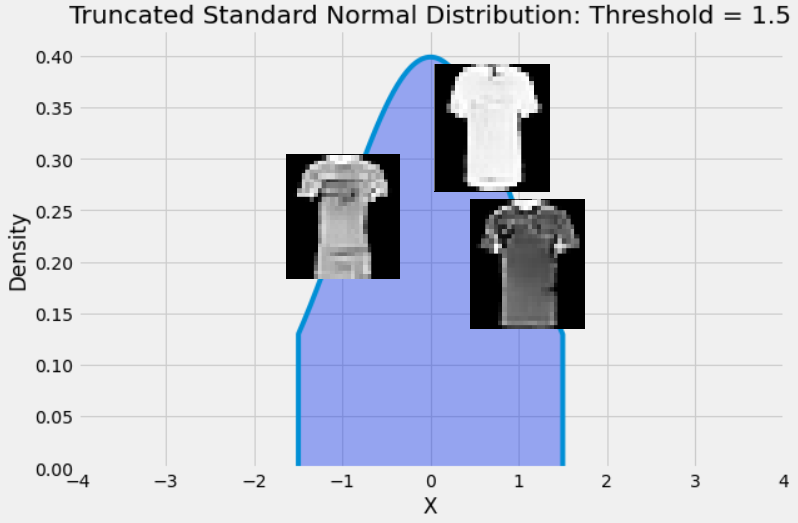}
         \caption{Truncated distribution}
     \end{subfigure}
     
     \caption{ Drawing the generator's latent vector $\emph{z}$ from a truncated normal distribution allows us to feed in the most realistic generated samples for classification. }
     
\end{figure}

  


%


\vspace{-3mm}

\section{Experiments}
\label{sec:experiments}

To examine the effectiveness of AC-GANs in classification, we conducted experiments on three datasets of different complexity. We utilized the Fashion-MNIST dataset, which consists of $28$x$28$ grayscale images of clothing items from $10$ classes~\cite{xiao2017fashion}. Additionally, we used the CIFAR10 dataset, a collection of $32$x$32$x$3$ natural images of 10 classes~\cite{krizhevsky2009learning}. To address higher dimensional spaces, we used the COVID-19 Radiography Database~\cite{rahman_2021}, resizing the chest X-rays to $128$x$128$x$3$ for COVID-positive versus negative classification. 

We first examined the utility of the modified AC-GAN scheme on different training set sizes in comparison to a standard CNN and a standard AC-GAN. The AC-GAN is modified with a gradient penalty Wasserstein loss and truncation, as detailed in Section 3. From here on out, we refer to the modified AC-GAN as WAC-GAN-GPT (Wasserstein AC-GAN with Gradient Penalty and Truncation). We trained a standard CNN classifier on Fashion-MNIST using training set sizes of 500, 2500, 10000, 20000, and 40000 training examples with random horizontal and vertical flips for augmentation. For each trained model, we used 5000 images for validation and evaluated on a held-out test set of 10000 images. Next, we constructed a standard AC-GAN. To maintain a controlled setting for a fair comparison between the CNN and AC-GAN for classification, we maintained the same feature extraction and classification layers. We fed in a generator into the baseline CNN architecture, and to facilitate the generator's training, we added a layer to the CNN for discriminating between real and fake images. The AC-GAN was similarly trained on the incremental training sets, but without any traditional augmentation, and evaluated on the same test set. Lastly, we modified the AC-GAN framework using the methods we proposed in Section~\ref{sec:method}. Each modification, the truncation trick and Wasserstein loss with gradient penalty, is independently evaluated through ablation. \\
\indent To examine the ability for the AC-GAN  to generalize from classifying on one data distribution to a slightly shifted distribution, we used $40000$ training and $10000$ validation images from the CIFAR10 dataset. We then used the so-called CIFAR 10.1v6 dataset for testing, which consists of $2000$ images from the Tiny Images dataset~\cite{torralba200880} with equal balance for each of the ten classes corresponding to CIFAR10. The authors of~\cite{recht2018cifar} constructed this dataset and empirically show that it follows a slightly varied distribution from CIFAR10. We trained an AlexNet architecture~\cite{krizhevsky2017imagenet} on the CIFAR10 training set as~\cite{recht2018cifar} notes the greatest drop in performance on CIFAR10 to CIFAR 10.1v6 for this architecture. We then used the AlexNet architecture as the discriminator for our AC-GAN scheme.

Finally, we trained a CNN (with traditional data augmentation), standard AC-GAN, and WAC-GAN-GPT on a sample of the COVID-19 Radiography database at a training size of 800 with even class split, then validated on 1000 images. All common hyperparameters and initialization schemes were shared between the trained CNNs and AC-GANs for all experiments. Architecture and training details are further specified at  \href{https://github.com/avdravid/AC-GANS-FOR-IMAGE-CLASSIFICATION}{https://github.com/avdravid/AC-GANS-FOR-IMAGE-CLASSIFICATION}.
%

\section{Results and Discussion}
\label{sec:results}
On the Fashion-MNIST dataset, the WAC-GAN-GPT outperforms both the baseline CNN model and standard AC-GAN by ${\sim}{1}{-}{5}\%$ based on all training set sizes, as detailed in Table 1 and Figure 2. The CNN was trained using traditional affine transformations for augmentation, but the classifier/discriminator in the AC-GAN frameworks were not. This demonstrates that the generated images provide more meaningful additional data to classify. Both the AC-GAN with truncation and AC-GAN with gradient-penalty Wasserstein loss (WAC-GAN-GP) show slight improvement over the CNN and standard AC-GAN, with the truncation providing a greater contribution to performance gain. The combination of the two (WAC-GAN-GPT) results in the most competitive results of the five tested frameworks. The standard AC-GAN has comparable performance but does not outperform the baseline CNN at all training set sizes. GANs need sufficient data to synthesize decent quality images~\cite{noguchi2019image}. If there is greater stochasticity in sampling the generator's images, especially when the AC-GAN is trained on a small dataset, the generated images fed into the discriminator may be of poor quality, which can hurt the classification training process. 

\begin{table}[h]
    
\resizebox{\linewidth}{!}{
\renewcommand{\tabcolsep}{5pt}
\begin{tabular}{c|c|c|c|c|c}
\textbf{Train Size} & \multicolumn{1}{c|}{\begin{tabular}[c]{@{}c@{}}Baseline \\ CNN\end{tabular}} & \multicolumn{1}{c|}{AC-GAN} & WAC-GAN-GP & \multicolumn{1}{c|}{\begin{tabular}[c]{@{}c@{}}AC-GAN \\ with Truncation\end{tabular}} & \multicolumn{1}{c}{\begin{tabular}[c]{@{}c@{}}WAC-GAN-GPT\end{tabular}} \\ \cline{2-6} 
500 &   $77.5\%\pm1.5$ &    $77.6\%\pm1.7$ &  $77.9\%\pm1.5$ & $78.8\%\pm1.5$ &  $\mathbf{79.8}\%\pm1.5$\\
2500 & $83.5\%\pm1.0$&    $81.2\%\pm2.1$ & $84.4\%\pm1.5$ & $84.8\%\pm 1.1$ & $\mathbf{86.0}\%\pm1.2$\\
10000       & $86.4\%\pm1.5$ & $87.3\%\pm1.3$   & $87.6\%\pm0.9$& $87.8\%\pm0.7$        & $\mathbf{88.4}\%\pm1.1$  \\
20000   &   $87.5\%\pm1.3$ &  $88.6\%\pm1.6$ & $88.1\%\pm1.2$  &  $89.1\%\pm0.5$
& $\mathbf{89.8}\%\pm0.9$\\
40000  & $90.3\%\pm0.8$ & $90.9\%\pm0.8$ & $91.0\%\pm0.4$  & $90.7\%\pm0.8$     
& $\mathbf{91.3}\%\pm0.7$                    
\end{tabular}

}
\captionof{table}{95\% confidence intervals for accuracy on Fashion MNIST test set as a function of training set size. Best results in bold.} 
\label{table1}
\end{table}

To examine the distributional relationships between the AC-GANs and CNN at the low data regime, we conduct t-sNE analysis~\cite{van2008visualizing} using the CNN, AC-GAN, and WAC-GAN-GPT trained on just 500 Fashion MNIST samples. We sample 300 real images, 300 generated images from the standard AC-GAN and 300 images drawn using a truncated standard normal distribution from the WAC-GAN-GPT. We then feed them through the CNN to obtain feature embeddings, subsequently applying t-SNE to transform the embeddings to a two-dimensional space and visualizing the data based on either the true label for real images or the label given to the generator for generated images. An example is shown in Figure 3. The average distance for each data point to the center of its class cluster is 7.83, 5.16, and 3.94 for the CNN, AC-GAN, and WAC-GAN-GPT respectively. The standard deviations for these distances is 4.71, 2.17, and 1.76, respectively. Multiple runs of t-SNE confirm these trends. Although greater diversity of training data is vital for improving generalizability, based on the higher performance of the WAC-GAN-GPT in the low data regime, we find that the samples closer to the mean are informative in these early stages of learning. The classifier needs to first learn the most common features following a simple distribution before learning more complex, diverse ones. The standard AC-GAN without truncation may produce confusing images to train a classifier, some of which may cross the decision boundaries, which may harm training.

\begin{figure}[]
\vspace{-.5cm}
\begin{minipage}[b]{1\linewidth}
  \centering
  \centerline{\includegraphics[width=7cm]{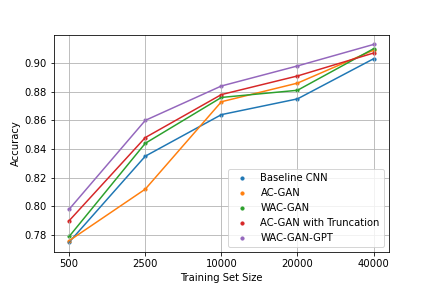}}
\end{minipage}
\caption{Graphical comparison of average accuracy on the test set for the five models as training set size increases. All methods show comparable performance. The WAC-GAN-GPT consistently improves over the baselines, especially in the lower data regime.}
\label{fig:fashion-mnistplot}
\end{figure}

\begin{figure}
  
  \centering

     \begin{subfigure}[t]{0.49\linewidth}
         \centering
         \includegraphics[width=\linewidth]{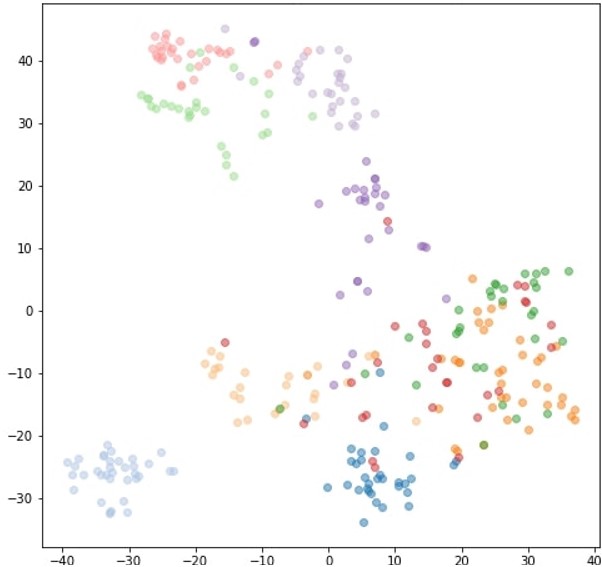}
         \caption{True Samples}
     \end{subfigure}
     \hfill
     \begin{subfigure}[t]{0.49\linewidth}
         \centering
         \includegraphics[width=\linewidth]{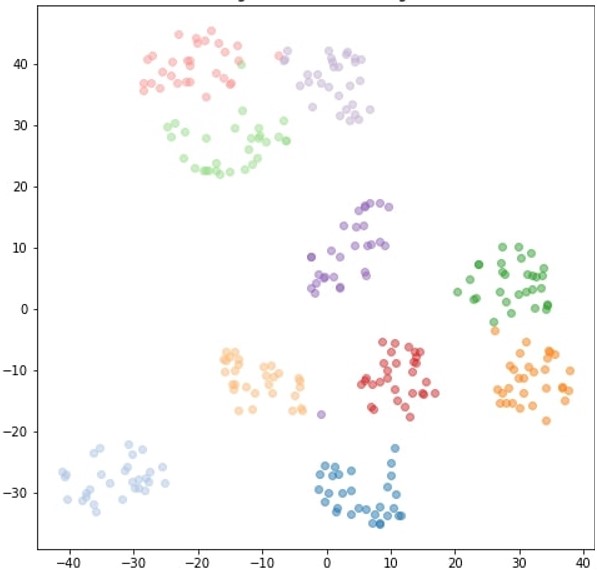}
         \caption{AC-GAN Samples}
     \end{subfigure}
     \hfill
     \begin{subfigure}[t]{0.49\linewidth}
         \centering
         \includegraphics[width=\linewidth]{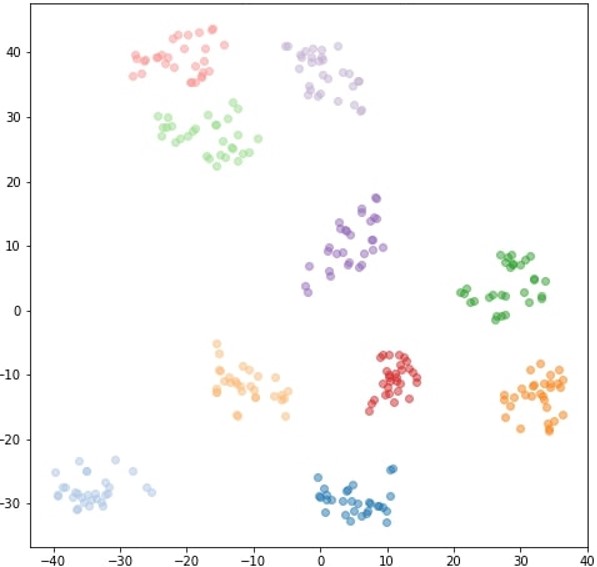}
         \caption{WAC-GAN-GPT Samples}
     \end{subfigure}
     
     \caption{ T-SNE of Fashion MNIST true and generated samples.  }
\end{figure}

On the CIFAR10 dataset, the WAC-GAN-GPT method outperforms the standard CNN with traditional augmentation and the standard AC-GAN on both the CIFAR10 and the CIFAR10.1v6 dataset, as seen in Table 2. The drop in absolute accuracy from CIFAR10 to CIFAR10.1v6 is explained by a natural distribution shift in this new test set~\cite{recht2018cifar}. The discrepancy in the different test set accuracies is reduced from ${\sim17\%}$ with the CNN to ${\sim14\%}$ with the WAC-GAN-GPT. This suggests a greater generalization ability from one dataset set to another for the WAC-GAN-GPT. We consider that this is due to the nature of \emph{implicit density estimation}: the generator's sampled images from the estimated $p_g$ distribution does not perfectly match $p_{data}$. As the discriminator trains its classification task on both $p_g$ and $p_{data}$, it does not overfit to $p_{data}$ and may be able to better adapt to slightly varied distributions. 

The potential for the AC-GAN for image classification is further corroborated with experiments in a higher resolution space: $128$x$128$x$3$ images from the COVID-19 Radiography Database. Results are seen in Table 2 above.

\begin{table}[]
    
\resizebox{\linewidth}{!}{
\vspace{.3cm}
\centering
\tiny
\begin{tabular}{cccc}
\multicolumn{1}{c|}{}             & AlexNet              & AC-GAN               & WAC-GAN-GPT          \\ \hline
\multicolumn{1}{c|}{CIFAR 10} & $70.5\%\pm 0.5$      & $70.1\%\pm 0.8$      & $\mathbf{72.9\%}\pm 0.7$ \\
\multicolumn{1}{c|}{CIFAR 10.1v6} & $53.5\%\pm 1.0$      & $56.4\%\pm 1.1$      & $\mathbf{59.3\%}\pm 0.6$ \\
\multicolumn{1}{l}{}              & \multicolumn{1}{l}{} & \multicolumn{1}{l}{} & \multicolumn{1}{l}{}     \\
\multicolumn{1}{c|}{}             & CNN                  & AC-GAN               & WAC-GAN-GPT          \\ \hline
\multicolumn{1}{c|}{COVID-19}     & $94.0\%\pm1.5$       & $95.5\%\pm0.5$       & $97.6\%\pm0.9$          
\end{tabular}
}
\captionof{table}{95\% confidence intervals for accuracy on CIFAR10/CIFAR10.1v6 test sets and the COVID-19 Radiography Database validation set.} 

\end{table}

\vspace{-3mm}

\section{Conclusion}
\label{sec:conclusion}
We have demonstrated that the AC-GAN, in fact, can achieve competitive performance with standard CNNs across datasets of varying complexity and resolution, with particular performance gains in lower data regimes. We have presented some methods which can be employed to improve accuracy and save efforts on training a separate GAN. We hope this work inspires further efforts to interface GANs with image classification as a 'one-stop-shop.' Future work can look into using AC-GANs with more diverse datasets, higher resolution images, and interfacing them with more advanced techniques, such as adaptive discriminator augmentation or progressive growing~\cite{karras2020training, karras2017progressive}. This could further facilitate training with limited data and generating higher fidelity images to potentially improve upon the current benchmarks in image classification.

{
\ninept
\balance
\bibliographystyle{IEEEbib}
\bibliography{references}
}

\end{document}